\documentclass[11pt,a4paper]{article}
\usepackage[hyperref]{emnlp-ijcnlp-2019}
\usepackage{times}
\usepackage{latexsym}

\usepackage{url}
\usepackage{stfloats}
\usepackage{url}
\usepackage{float}
\usepackage{graphicx}
\usepackage{epstopdf}
\usepackage{amsfonts}
\usepackage{amssymb}
\usepackage{amsmath}
\usepackage{verbatim}
\usepackage{multirow}
\usepackage{amsmath}
\usepackage[noend]{algorithmic}
\usepackage[ruled]{algorithm2e}
\usepackage{subfigure}
\usepackage{pifont}
\usepackage{array}
\usepackage{fp}
\usepackage{makecell}
\usepackage{flushend}
\usepackage{cases}
\usepackage{color}
\usepackage{booktabs}
\usepackage{bbm}
\usepackage[ruled]{algorithm2e}
\usepackage{multirow}

\usepackage{CJK}

\aclfinalcopy 

\newcommand{\ignore}[1]{}

\title{Unsupervised Context Rewriting for Open Domain Conversation}

\author{
	Kun Zhou\textsuperscript{\rm{1\thanks{\ \ This work was done when the first author was an intern at Microsoft XiaoIce.}}}, 
	Kai Zhang\textsuperscript{\rm{2}}, 
	Yu Wu\textsuperscript{\rm{3}},
	Shujie Liu\textsuperscript{\rm{3}},
	{\rm and} Jingsong Yu\textsuperscript{\rm{1}} \\
	\textsuperscript{1}School of Software and Microelectronics, Peking University \\
	\textsuperscript{2}AI and Research Microsoft, Beijing, China\ \\
	\textsuperscript{3}Microsoft Research, Beijing, China\\
	\texttt{franciszhou@pku.edu.cn} \\
	\texttt{\{kaizh,Wu.Yu,shujliu\}@microsoft.com} \\ \texttt{yjs@ss.pku.edu.cn}\\
}

\date{}
\begin{document}
\maketitle
\begin{abstract}
  Context modeling has a pivotal role in open domain conversation. Existing works either use heuristic methods or jointly learn context modeling and response generation with an encoder-decoder framework. This paper proposes an explicit context rewriting method, which rewrites the last utterance by considering context history. We leverage pseudo-parallel data and elaborate a context rewriting network, which is built upon the CopyNet with the reinforcement learning method. The rewritten utterance is beneficial to candidate retrieval, explainable context modeling, as well as enabling to employ a single-turn framework to the multi-turn scenario. The empirical results show that our model outperforms baselines in terms of the rewriting quality, the multi-turn response generation, and the end-to-end retrieval-based chatbots. 
\end{abstract}

\section{Introduction}
Recent years have witnessed remarkable progress in open domain conversation (non-task oriented dialogue system) \cite{ji2014information, li2015diversity} due to the easy-accessible conversational data and the development of deep learning techniques \cite{bahdanau2014neural}.  One of the most difficult problems for open domain conversation is how to model the conversation context. 
	
A conversation context is composed of multiple utterances, which raises some challenges not existing in the sentence modeling, including: 1) topic transition; 2) plenty of coreferences (he, him, she, it, they); and 3) long term dependency. To tackle these problems, existing works either refine the context by appending keywords to the last turn utterance \cite{DBLP:conf/sigir/YanSW16}, or learn a vector representation with neural networks \cite{serban2017hierarchical}. However, these methods have drawbacks, for instance, correct keywords cannot be selected by heuristics rules, and a fix-length vector is not able to handle a long context. 
	
\begin{figure}[t!]
    \centering
    \includegraphics[width=.98\linewidth]{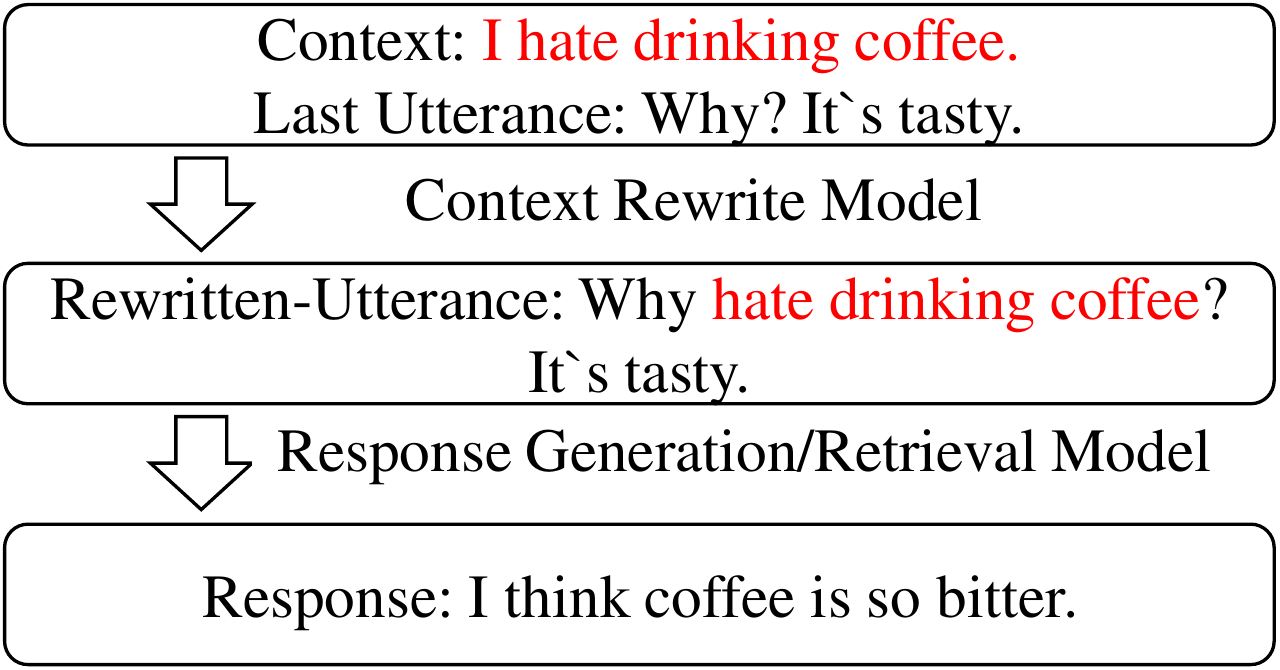}
    \caption{An Example of Context Rewriting.    \label{fig1}}\vspace{-3mm}
    \end{figure}

We propose a context rewriting method, which explicitly rewrites the last utterance by considering the contextual information. Our goal is to generate a self-contained utterance, which neither has coreferences nor depends on other utterances in history. By this means, we change the input of chatbots from an entire conversation session to a rewritten sentence, which significantly reduces the difficulty of response generation/selection since the rewritten sentence is shorter and does not has redundant information. Figure \ref{fig1} gives an example to further illustrate our idea.

The last utterance contains the word ``it" which refers to the coffee in context. Moreover, ``Why?" is an elliptical interrogative sentence, which is a shorter form of ``Why hate drinking coffee?". We rewrite the context and yield a self-contained utterance ``Why hate drinking coffee? It's tasty."  Compared to previous methods, our method enjoys the following advantages: 1) The rewriting process is friendly to the retrieval stage of retrieval-based chatbots. Retrieval-based chatbots consists of two components: candidate retrieval and candidate reranking. Traditional works \cite{DBLP:conf/sigir/YanSW16,wu2016sequential} pay little attention to the retrieval stage, which regards the entire context or context rewritten with heuristic rules as queries so noise is likely to be introduced;  2) It makes a step toward explainable and controllable context modeling, because the explicit context rewriting results are easy to debug and analyze. 3) Rewritten results enable us to employ a single-turn framework to solve the multi-turn conversation task. The single-turn conversation technology is more mature than the multi-turn conversation technology, which is able to achieve higher responding accuracy.

To this end, we propose a context rewriting network (CRN) to integrate the key information of the context and the original last utterance to build a rewritten one, so as to improve the answer performance.
Our CRN model is a sequence-to-sequence network \cite{sutskever2014sequence} with a bidirectional GRU-based encoder, and a GRU-based decoder enhanced with the CopyNet \cite{DBLP:conf/acl/GuLLL16}, which helps the CRN to directly copy words from the context.
Due to the absence of the real written last utterance, unsupervised methods are used with two training stages, a pre-training stage with pseudo rewritten data, and a fine-tuning stage using reinforcement learning (RL) \cite{sutton1998introduction} to maximize the reward of the final answer. Without the pre-training part, RL is unstable and slow to converge, since the randomly initialized CRN model cannot generate reasonable rewritten last utterance. On the other hand, only the pre-training part is not enough, since the pseudo data may contain errors and noise, which restricts the performance of our CRN.

We evaluate our method with four tasks, including the rewriting quality, the multi-turn response generation, the multi-turn response selection, and the end-to-end retrieval-based chatbots. Empirical results show that the outputs of our method are closer to human references than baselines. Besides, the rewriting process is beneficial to the end-to-end retrieval-based chatbots and the multi-turn response generation, and it shows slightly positive effect on the response selection.
	

\section{Related Work}
Recently, data-driven approaches for chatbots  \cite{ritter2011data,ji2014information} has achieved promising results. Existing work along this line includes retrieval-based methods \cite{hu2014convolutional,ji2014information,wang2015syntax,DBLP:conf/sigir/YanSW16,zhou2016multi} and generation-based methods \cite{DBLP:conf/acl/ShangLL15,serban2015building,vinyals2015neural,li2015diversity,li2016persona,xing2016topic,serban2016multiresolution}.

Early research into retrieval-based chatbots \cite{wang2013dataset,hu2014convolutional,wang2015syntax} only considers the last utterances and ignores previous ones, which is also called Short Text Conversation (STC). Recently, several studies  \cite{lowe2015ubuntu,DBLP:conf/sigir/YanSW16,wu2016sequential,Wu2018ResponseGB}  have investigated multi-turn response selection, and obtained better results in a comparison with STC.  A common practice for multi-turn retrieval-based chatbots first retrieve candidates from a large index with a heuristic context rewriting method. For example, \cite{wu2016sequential} and \cite{DBLP:conf/sigir/YanSW16}  refine the last utterance by appending keywords in history, and retrieve candidates with the refined utterance. 
Then, response selection methods are applied to measure the relevance between history and candidates.
    
A number of studies about generation-based chatbots have considered multi-turn response generation. \newcite{sordoni2015neural} is the pioneer of this type of research, it encodes history information into a vector and feeds to the decoder. \newcite{DBLP:conf/acl/ShangLL15} propose three types of attention to utilize the context information. In addition, \newcite{serban2015building} propose a Hierarchical Recurrent Encoder-Decoder model (HRED), which employs a hierarchical structure to represent the context. After that, latent variables \cite{serban2017hierarchical} and hierarchical attention mechanism \cite{xing2017hierarchical} have been introduced to modify the architecture of HRED. Compared to previous work, the originality of this study is that it proposes a principle way instead of heuristic rules for context rewriting, and it does not depend on parallel data. \cite{Su_2019} propose an utterance rewrite approach but need human annotation.

\section{Model}
Given a dialogue data set \(\mathcal{D}=\{(U,r)_{z}\}^{N}_{z=1}\), where \(U=\{u_{0},\cdots,u_{n}\}\) represents a sequence of utterances and $r$ is a response candidate.
We denote the last utterance as $q = u_{n}$ for simplicity, which is especially important to produce the response, and other utterances as $c = \{u_{0},\cdots,u_{n-1}\}$. The goal of our paper is to rewrite $q$ as a self-contained utterance $q^{*}$ using useful information from $c$, which can not only reduce the noise in multi-turn context but also leverage a more simple single-turn framework to solve the multi-turn end-to-end tasks. We focus on the multi-turn response generation and selection tasks.

To rewrite the last utterance $q$ with the help of context $c$, we propose a context rewriting network (CRN), which is a popularly used sequence-to-sequence network, equipped with a CopyNet to copy words from the original context $c$ (Section \ref{Context Rewriting Network}). 
Without the real paired data (pairs of the original and rewritten last utterance), our CRN model is firstly pre-trained with the pseudo data, generated by inserting extracted keywords from context into the original last utterance $q$ (Section \ref{Pre-training with Pseudo Data}). To let the final response to influence the rewriting process, reinforcement learning is leveraged to further enhance our CRN model, using the rewards from the response generation and selection tasks respectively (Section \ref{Fine-Tuning with Reinforcement Learning}). 

\begin{figure*}[t]
\centering
\setlength{\abovecaptionskip}{0.3cm}
\setlength{\belowcaptionskip}{-0.5cm} 
\includegraphics[width=.8\linewidth]{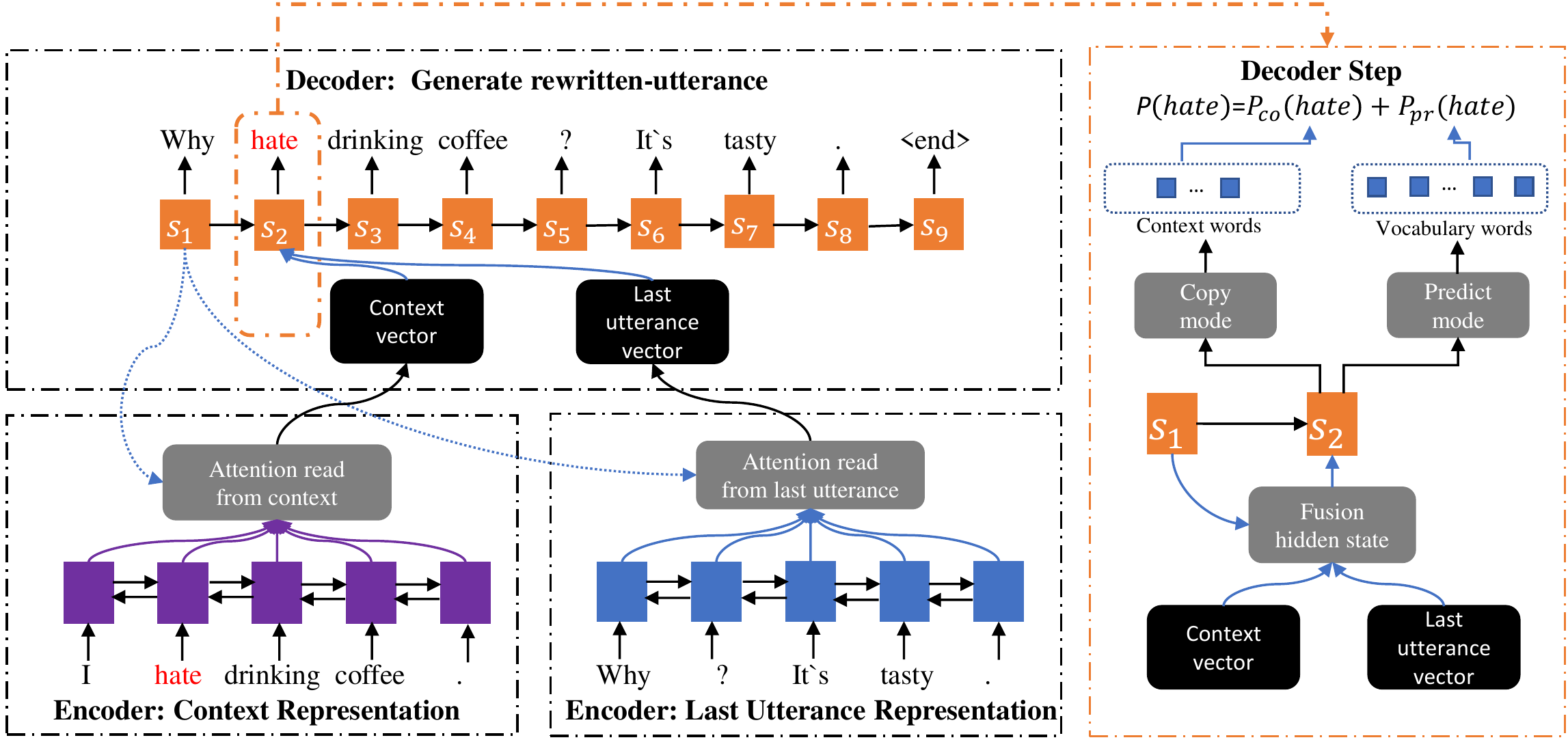}
\caption{The Detail of CRN    \label{fig2}}
\end{figure*}

\subsection{Context Rewriting Network}
\label{Context Rewriting Network}
As shown in Figure 2, our context rewriting network (CRN) follows the sequence to sequence framework, consisting of three parts: one encoder to learn the context ($c$) representation, another encoder to learn the last utterance ($q$) representation, and a decoder to generate the rewritten utterance $q^{*}$.  Attention is also used to focus on different words in the last utterance $q$ and the context $c$, and the copy mechanism is introduced to copy important words from the context $c$.

\subsubsection{Encoder}
\label{Encoder}
To encode the context $c$ and the last utterance $q$, bidirectional GRU is leveraged to 
take both the left and right words in the sentence into consideration, by concatenating the hidden states of two GRU networks in positive time direction and negative time direction. With the bidirectional GRU, the last utterance $q$ is encoding into $H_{Q}=[h_{q_{1}},\dots,h_{q_{nq}}]$, and the context $c$ is encoding into $H_{C}=[h_{c_{1}},\dots,h_{c_{nc}}]$.

\subsubsection{Decoder}
\label{Decoder}
The GRU network is also leveraged as decoder to generate the rewritten utterance $q^{*}$, in which the attention mechanism is used to extract useful information from the context $c$ and the last utterance $q$, and the copy mechanism is leveraged to directly copy words from the context $c$ into $q^{*}$. At each time step $t$, we fuse the information from $c$, $q$ and last hidden state $s_t$ to generate the input vector $z_t$ of GRU as following
\begin{equation}
\setlength{\abovedisplayskip}{5pt}
\setlength{\belowdisplayskip}{5pt}
    z_{t}=W^{T}_{f}[s_{t};\sum_{i=1}^{nq}\alpha_{q_{i}}{h_{q_{i}}};\sum_{i=1}^{nc}\alpha_{c_{i}}{h_{c_{i}}}]+b 
\end{equation}
where [;] is the concatenation operation. $W_{f}$ and $b$ are trainable parameters, $s_{t}$ is the last hidden state of decoder GRU in step $t$,  $\alpha_{q}$ and $\alpha_{c}$ are the weights of the words in $q$ and $c$ respectively, derived by the attention mechanism as following
\begin{equation}
\setlength{\abovedisplayskip}{5pt}
\setlength{\belowdisplayskip}{1pt}
    \alpha_{i}=\frac{exp(e_{i})}{\sum_{j=1}^{n}exp(e_{j})} 
\end{equation}
\begin{equation}
\setlength{\abovedisplayskip}{1pt}
\setlength{\belowdisplayskip}{1pt}
    e_{i}=h_{i}W_{a}s_{t}
\end{equation}
where $h_{i}$ is the encoder hidden state of the $i^{th}$ word in $q$ or $c$, $W_{a}$ is the trainable parameter. 

The copy mechanism is used to predict the next target word according to the probability of $p(y_{t}|s_{t},H_{Q},H_{C})$, which is computed as 
\begin{equation}
\setlength{\abovedisplayskip}{5pt}
\setlength{\belowdisplayskip}{5pt}
\begin{split}
p(y_{t}|s_{t},H_{Q},H_{C})&=p_{pr}(y_{t}|z_{t}) \cdot p_{m}(pr|z_{t}) \\ 
&+p_{co}(y_{t}|z_{t}) \cdot p_{m}(co|z_{t})
\end{split}
\end{equation}
where $y_{t}$ is the $t-th$ word in response, $pr$ and $co$ stand for the predict-mode and the copy-mode, $p_{pr}(y_{t}|z_{t})$ and $p_{co}(y_{t}|z_{t})$ are the distributions of vocabulary word and context word which are implemented by two MLP (multi layer perceptron) classifiers, respectively. And $p_{m}(\cdot|\cdot)$ indicates the probability to choose the two modes, which is a MLP (multi layer perceptron) classifier with softmax as the activation function:
\begin{equation}
\setlength{\abovedisplayskip}{5pt}
\setlength{\belowdisplayskip}{5pt}
    p_{m}(pr|z_{t})=\frac{e^{\psi_{pr}(y_{t},H_{Q},H_{C})}}{e^{\psi_{pr}(y_{t},H_{Q},H_{C})}+e^{\psi_{co}(y_{t},H_{Q},H_{C})}}
\end{equation}
where $\psi_{pr}(\cdot)$, $\psi_{co}(\cdot)$ are score functions for choosing the predict-mode and copy-mode with different parameters. 

\subsection{Pre-training with Pseudo Data}
\label{Pre-training with Pseudo Data}
Instead of directly leverage RL to optimize our model CRN, which could be unstable and slow to converge, we pre-train our model CRN with pseudo-parallel data. Cross-entropy is selected as the training loss to maximize the log-likelihood of the pseudo rewritten utterance. 
\begin{equation}
\setlength{\abovedisplayskip}{5pt}
\setlength{\belowdisplayskip}{5pt}
    L_{MLE}=-\frac{1}{N}\sum_{i=1}^{n}\text{log}(p(y_{t}|s_{t},H_{Q},H_{C})) 
\end{equation}
The main challenge for the pre-training stage is how to generate good pseudo data, which can integrate suitable keywords from context and the original last utterance to form a better one to generate a good response. Given the context $c$, we extract keywords $w^{*}_{c_{1:n}}$ using pointwise mutual information (PMI) (in Section \ref{Key Words Extraction}). With the extracted keywords $w^{*}_{c_{1:n}}$, language model is leveraged to find suitable positions to insert them into the original last utterance to generate rewritten candidates $s^{*}$, which will be re-ranked leveraging the information from following process (the response generation/selection) to get final pseudo rewritten utterance $Q^{*}$ (in Section \ref{Pseudo Data Generation}). In the following of this section, we will introduce our pseudo data creation method in detail.



\subsubsection{Key Words Extraction}
\label{Key Words Extraction}
To penalize common and low frequent words, and prefers the ``mutually informative'' words, PMI is used to extract the keywords in the context $c$. Given a context word $w_c$, and a word $w_r$ in response $r$, it is the divides the prior distribution $p_{c}(w_{c})$ by the posterior distribution $p(w_{c}|w_{r})$ as shown as:
\begin{equation}
\setlength{\abovedisplayskip}{5pt}
\setlength{\belowdisplayskip}{5pt}
    \text{PMI}(w_{c},w_{r})=-\text{log}\frac{p_{c}(w_{c})}{p(w_{c}|w_{r})}
\end{equation}
In order to select the keywords which contribute to the response, and are suitable to be shown in the last utterance, we also calculate $\text{PMI}(w_{c},w_{q})$ between the context word $w_c$ and any word $w_q$ in the last utterance. 
The final contribution score $\text{PMI}(w_{c},q,r)$ for the context word $w_c$ to the last utterance $q$ and the response $r$ is calculated as
\begin{equation}
\setlength{\abovedisplayskip}{5pt}
\setlength{\belowdisplayskip}{5pt}
\text{norm}(\text{PMI}(w_{c}, q)) +\text{norm}(\text{PMI}(w_{c}, r))
\end{equation}
where norm$(\cdot)$ is the min-max normalization among all words in $c$, and $\text{PMI}(w_{c},q)$ (similar for $\text{PMI}(w_{c},r)$) is calculated as  
\begin{equation}
\setlength{\abovedisplayskip}{5pt}
\setlength{\belowdisplayskip}{5pt}
\text{PMI}(w_{c}, q)=\sum_{w_q \in q}\text{PMI}(w_{c},w_{q}).
\end{equation}
The keywords $w^{*}_{c}$ with top-$20\%$ contribution score $\text{PMI}(w_{c},q,r)$ against $r$ and $q$ are selected to insert into the last utterance $q$.\footnote{This threshold of PMI is based on the observation on the development set.}

\subsubsection{Pseudo Data Generation}
\label{Pseudo Data Generation}
Together with the extract candidate keyword, the words nearby are also extracted to form a continuous span to introduce more information, of which, at most 2 words before and after are considered. For one keyword, there are at most $C_{3}^{1}*C_{3}^{1}=9$ span candidates. We apply a multi-layer RNN language model to insert the extracted key phrase to a suitable position in the last utterance. Top-3 generated sentences with high language model scores are selected as the rewritten candidates $s^{*}$. 

With the information from the response, a re-rank model is used to select the best one from the candidates $s^{*}$. For end-to-end generation task, the quality of candidates is measured with the cross-entropy of a single-turn attention based encoder-decoder model $M_{s2s}$, hoping that the good rewritten utterance can help to generate the proper response. 
For the end-to-end response selection task, the quality of the candidates is measured by the rank loss of a single-turn response selection model $M_{ir}$, hoping that the good one can distinguish the positive and negative responses. 
\begin{equation}
\label{s2s loss}
\setlength{\abovedisplayskip}{1pt}
\setlength{\belowdisplayskip}{1pt}
    L_{M_{s2s}}(r|s^{*})=-\frac{1}{n}\sum_{i=1}^{n}\log p(r_{1},\dots,r_{n}|s^{*})
\end{equation}

\begin{equation}
\label{ir loss}
\setlength{\abovedisplayskip}{-5pt}
\setlength{\belowdisplayskip}{1pt}
    L_{M_{ir}}(po,ne,s^{*})=M_{ir}(po,s^{*})-M_{ir}(ne,s^{*})
\end{equation}
In equation~\ref{s2s loss}, $r_{i}$ is the $i-th$ word in response. In equation~\ref{ir loss} $po$ is the positive response, and $ne$ are the negative one.


\subsection{Fine-Tuning with Reinforcement Learning}
\label{Fine-Tuning with Reinforcement Learning}
Since the generated pseudo data inevitably contains errors and noise, which limits the performance of the pre-trained model, we leverage the reinforcement learning method to build the direct connection between the context rewrite model CRN and different tasks. We first generate rewritten utterance candidates $q_{r}$ with our pre-trained model, and calculate the reward $R(q_{r})$ which will be maximized to optimize the network parameters of our CRN. Due to the discrete choices of words in sequential generation, the policy gradient is used to calculate the gradient. 
\begin{equation}
\setlength{\abovedisplayskip}{5pt}
\setlength{\belowdisplayskip}{5pt}
    \nabla_{\theta}J(\theta)=E[R\cdot\nabla \log(P(y_{t}|x))]
\end{equation}
For reinforcement learning in sequential generation task, instability is a serious problem. Similar to other works \cite{wu2018reinforce}, we combine MLE training objective with RL objective as
\begin{equation}
\setlength{\abovedisplayskip}{5pt}
\setlength{\belowdisplayskip}{5pt}
    L_{com}=L_{rl}^{*}+\lambda L_{MLE}
\end{equation}
where $\lambda$ is a harmonic weight.

By directly maximizing the reward from end tasks (response generation and selection), we hope that our CRN can correct the errors in the pseudo data and generate better rewritten last utterance. Two different rewards are used to fine-tune our CRN for the tasks of response generation and selection respectively. We will introduce them in detail in the following.

\subsubsection{End-to-end response generation reward}
Similar as we do in Section \ref{Pseudo Data Generation},  for end-to-end response generation task, we use the cross-entropy loss of a single-turn attention based encoder-decoder model $M_{s2s}$ to evaluate the quality of rewritten last utterance $q_{r}$ as
\begin{equation}
\label{reward for generation}
\setlength{\abovedisplayskip}{5pt}
\setlength{\belowdisplayskip}{1pt}
    R_{g}(r,q^{*},q_{r})=L_{M_{s2s}}(r|q^{*})-L_{M_{s2s}}(r|q_{r})
\end{equation}
where $L_{M_{s2s}}$ is defined in Equation \ref{s2s loss}, $r$ is the response candidate, $q_{r}$ is the generated candidate of our CRN, and $q^{*}$ is the pseudo rewritten candidate as introduced in Section \ref{Pseudo Data Generation}. If $q_{r}$ can bring more useful information from context, it will get lower cross-entropy to generate $r$ than the original pseudo rewritten one $q^{*}$. 

\subsubsection{End-to-end response selection reward}
\label{selection reward}
For end-to-end response selection task, we use a single-turn response selected model $M_{ir}$ to evaluate the quality of the generated candidate $q_{r}$ by the rank loss, it is calculated as
\begin{equation}
\setlength{\abovedisplayskip}{5pt}
\setlength{\belowdisplayskip}{5pt}
\begin{split}
    R_{ir}(po,ne,q^{*},q_{r})&=L_{M_{ir}}(po,ne,q_{r}) \\
    &-L_{M_{ir}}(po,ne,q^{*})
\end{split}
\end{equation}
where $L_{M_{ir}}$ is defined in Equation \ref{ir loss}, $q_{r}$ is the generated candidate of our model, and $q^{*}$ is the pseudo candidate as introduced in Section \ref{Pseudo Data Generation}. Similar to Equation \ref{reward for generation}, if $q_{r}$ can bring more useful information, it will do better to distinguish the negative and positive responses.

\section{Experiment}
We conduct four experiments to validate the effectiveness of our model, including the rewriting quality, the multi-turn response generation, the multi-turn response selection, and the end-to-end retrieval-based chatbots.

We crawl human-human context-response pairs from Douban Group which is a popular forum in China and remove duplicated pairs and utterances longer than 30 words. We create pseudo rewritten context as described in Section \ref{Context Rewriting Network}. Because most of the responses are only relevant with the last two turn utterances, following \newcite{li2016deep}, we remove the utterances beyond the last two turns. We finally split 6,844,393 ($c_{i},q_{i},q^{*}_{i},r_{i}$) quadruplets for training\footnote{The data in the training set do not overlap with the test data of the four tasks.}, 1000 for validation and 1074 for testing, and the last utterance in test set are selected by human and they all require rewriting to enhance information\footnote{Our human-annotated test set is available at \url{https://github.com/love1life/chat}}. In the data set, the ratio between rewritten last utterance and un-rewritten\footnote{Un-rewritten ones can handle utterances do not rely on their contexts.} the last utterance is $1.426:1$. The average length of context $c_{i}$, last utterance $q_{i}$, response $r_{i}$, and rewritten last utterance $q^{*}_{i}$ are 12.69, 11.90, 15.15 and 14.27 respectively.

We pre-train the CRN with the pseudo-parallel data until it coverages, then we use the reinforcement learning technique described in Section \ref{Fine-Tuning with Reinforcement Learning} to fine-tune the CRN. The specific details of the model hyper-parameters and optimization algorithms can be found in the \textbf{Supplementary}.


\subsection{Rewriting Quality Evaluation}
\label{quality_eval}
The detail of the training process is the same as Section \ref{Implementation Details0}. We evaluate the rewriting quality by calculating the BLEU-4 score \cite{papineni2002bleu}, a sequence order sensitive metric, between the system outputs and human references. Such references are rewritten by a native speaker who considers the information in context. It is required that the rewritten last utterance is self-contained. 

\begin{table}[t]

			\centering
			\setlength{\abovecaptionskip}{0.2cm}
			\setlength{\belowcaptionskip}{-0.5cm} 
	    	\small
			
			\begin{tabular}{l|c}
				\toprule
				& BLEU-4  \\ \midrule
				
				Last Utterance &34.2\\ 
				Last Utterance + Context &37.1 \\ 
				Last Utterance + Keyword &49.8 \\ 	 \midrule		 		
				CRN & 50.9 \\ 
				CRN + RL & \textbf{54.2} \\
				\bottomrule
				
			\end{tabular}	
					\caption{The result of rewriting quality.	\label{exp:1}  }	
		\end{table}
		
We compare our models CRN with three baselines. Firstly, we report the BLEU-4 scores of the origin last utterance and the combination of the last utterance and context. Additionally, following  \newcite{wu2016sequential}, we append five keywords to the last utterance, where the keywords are selected from the context by TF-IDF weighting, which is named by Last Utterance + Keyword. The IDF score is computed on the entire training corpus. 

Table \ref{exp:1} shows the experiment result, which indicates that our rewriting method outperforms heuristic methods. Moreover, a $54.2$ BLEU-4 score means that the rewritten sentences are very similar to the human references. CRN-RL has a higher score than CRN-Pre-train on BLEU-4, it proves reinforcement learning promotes our model effectively.

	\begin{table*}[t]
			\small
			\centering
			\begin{tabular}{l|c|c|c|c|c|c|c|c}
				\toprule
				& BLEU-1 & BLEU-2 & BLEU-3&  Average&  Extrema &  Greedy &  Distinct-1 &  Distinct-2\\ \midrule
				
				S2SA &5.72 &2.80 &1.37 &11.14 &8.58 &13.15 &25.55 &58.89\\ 
				HRED &10.10 &5.53 &2.75 &27.45 &21.71 &27.43 &15.22 &32.19\\ 
				Dynamic &7.05 &3.54 &1.75 &17.77 &14.20 &18.94 &6.22 &15.08\\
				Static &9.31 &5.01 &2.77 &21.32 &17.36 &22.49 &6.59 &17.31\\ \midrule
				CRN &13.26 &8.43 &4.64 &32.31 &26.97 &33.59 &\textbf{31.48} &\textbf{67.02} \\ 
				CRN + RL &\textbf{13.63} &\textbf{8.69} &\textbf{4.88} &\textbf{33.14} &\textbf{27.49} &\textbf{34.68} &31.42 &65.10 \\
				
				\bottomrule
			\end{tabular}
					\caption{Automatic evaluation results. \label{exp:gen} 	
			}		
		\end{table*}
		
\begin{table*}[t]

			\centering
			\setlength{\abovecaptionskip}{0.2cm}
			\setlength{\belowcaptionskip}{-0.5cm} 
			\small
			\begin{tabular}{l|c|c|c|c|c|c}
				\toprule
				& 3& 2& 1& 0& -1& avg\\ \midrule
				
				S2SA &14.48\% &\textbf{48.56}\% &31.01\% &5.48\% &0.46\% &1.72 \\
				HRED &13.28\% &16.90\% &\textbf{65.00}\% &3.99\% &0.46\% &1.40\\ \midrule
				CRN+S2SA &37.05\% &31.01\% &25.07\% &6.04\% &0.46\% &1.99\\
				CRN+S2SA+RL &\textbf{42.43}\% &29.25\% &15.04\% &\textbf{12.63}\% &0.46\% &\textbf{2.02}\\ \bottomrule

			\end{tabular}	
					\caption{The distribution of human evaluation in response generation model.	\label{exp:gen_human}  }	
		\end{table*}
\subsection{Multi-turn Response Generation}
\label{Multi-turn Response Generation}
Section \ref{quality_eval} demonstrates the outputs of our model are more similar to the human rewritten references. In this part, we will show the influence of the context rewriting for response generation. 

We use the same test data in Section \ref{quality_eval} to evaluate our model in the multi-turn response generation task. The multi-turn response generation is defined as, given an entire context consisting of multiple utterances, a model should generate informative, relevant, and fluent responses. We compare against the following previous works:

\textbf{S2SA:} We adopt the well-known Seq2Seq with attention \cite{bahdanau2014neural} model to generate responses by feeding the last utterances $q$ as source sentences. 

\textbf{HRED:}  \newcite{serban2015building} propose using a hierarchical encoder-decoder model to handle the multi-turn response generation problem, where each utterance and the entire session are represented by different networks.  

\textbf{Dynamic, Static:} \newcite{zhang2018} propose two state-of-the-art hierarchical recurrent attention networks for response generation. The dynamic model dynamically weights utterances in the decoding process, while the static model weights utterances before the decoding process.

\subsubsection{Implementation Details}
\label{Implementation Details0}

Given a context $c$ and last utterance $q$, we first rewrite them with the CRN. Then the rewritten last utterance $q'$ is fed to a single-turn generation model. The details of the model can be found in \textbf{Supplementary} and we set the same sizes of hidden states and embedding in all models. We regard two adjacent utterances in our training data to construct the training dataset (5,591,794 utterance-response pairs) for the single-turn generation model. We do not use the rewritten context as the input in the training phase, since we would like to guarantee the gain only comes from the rewriting mechanism at the inference stage.

\subsubsection{Evaluation Metrics}
\label{Evaluation Metrics}
We regard the human response as the ground truth, and use the following metrics:
 
\textbf{Word overlap based metrics}: We report BLEU score \cite{papineni2002bleu} between model outputs and human references. 

\textbf{Embedding based metrics}: As BLEU is not correlated with the human annotation perfectly, following \cite{DBLP:conf/emnlp/LiuLSNCP16}, we employ embedding based metrics, Embedding Average (Average), Embedding Extrema (Extrema), and Embedding Greedy (Greedy)  to evaluate results. The word2vec is trained on the training data set, whose dimension is $200$. 

\textbf{Diversity}: We evaluate the response diversity based on the ratios of distinct unigrams and bigrams in generated responses, denoted as Distinct-1 and Distinct-2 \cite{li2015diversity}.

\textbf{Human Annotation}: We ask three native speakers to annotate the quality of generated responses. We compare the quality of our model with HRED and S2SA. We conduct 5-scale rating: +3, +2, +1, 0 and -1. +3: the response is natural, informative and relevant with context; +2: the response is natural, informative, but might not be relevant enough; +1: the response is natural, but might not be informative and relevant enough (e.g., I don't know); 0: The response makes no sense, irrelevant, or grammatically broken; -1: The response or utterances cannot be understood.

\subsubsection{Evaluation Results}
\label{Evaluation Results}
Table \ref{exp:gen} presents the automatic evaluation results, showing that our models outperform baselines on relevance and diversity. Table \ref{exp:gen_human} gives the human annotation results, which also demonstrates the superiority of our models. Our models significantly improve response diversity, mainly because the rewritten sentence contains rich information that is capable of guiding the model to generate a specific output. After reinforcement learning our model promotes on BLEU and embedding metrics, it is because reinforcement learning can build the connection between the utterance-rewrite model with the response generation model for exploring better rewritten-utterance. But our model drops a little on the diversity metrics after reinforcement learning, this owes to the fact that the reward is biased to relevance rather than diversity. The similar phenomenon can be observed in the comparison of HRED and S2SA, which means that although relevance can increase by considering context information, general responses become more frequently concurrently.

Table \ref{exp:gen_human} presents the distribution of score in human evaluation, we can observe that most of the responses generated by HRED and S2SA get 1 or 2 in human evaluation, while most of the responses generated by our model can get 2 or 3. It proves that our model can reduce noisy from context and construct an informative utterance to generate high-quality response. However, after reinforcement learning our model gets more 0 and 3 score, that is because after reinforcement learning, our model becomes unstable and prefers to extract more words from context. The score of one candidate will increase or decrease a lot if useful keywords or wrong keywords are inserted into the last utterance, respectively. In fact, more utterances are rewritten better after reinforcement learning so the average evaluation score improves.

\subsection{Multi-turn Response Selection}
\label{Multi-turn Response Selection}
We also evaluate the multi-turn response selection task of retrieval-based chatbots, which aims to select proper responses from a candidate pool by considering the context.  We use the Douban Conversation Corpus released by \newcite{wu2016sequential}, which is created by crawling a popular Chinese forum, the Douban Group \footnote{\url{https://www.douban.com/group/explore}}, covering various topics. Its training set contains 0.5 million conversational sessions, and the validation set contains 50,000 sessions. The negative instances in both sets are randomly sampling with a 1:1 positive-negative ratio. The test set contains 1000 conversation contexts, and each context has 10 response candidates with human annotations.

	\begin{table*}[t]
		\small
		\centering
		\begin{tabular}{l|c|c|c|c|c|c}
			\toprule
			& MAP&MRR&P@1&  R$_{10}$@1 & R$_{10}$@2  & R$_{10}$@5\\ \midrule
			TF-IDF \cite{lowe2015ubuntu}   & 0.331 &0.359 &0.180 & 0.096&0.172& 0.405 \\ 
			RNN \cite{lowe2015ubuntu}  & 0.390 &0.422 &0.208&0.118&0.223&0.589\\ 
			CNN  \cite{lowe2015ubuntu}  & 0.417 &0.440 &0.226&0.121&0.252&0.647\\ 
			LSTM \cite{lowe2015ubuntu}  & 0.485 & 0.527 &0.320&0.187&0.343&0.720\\ 
			BiLSTM \cite{lowe2015ubuntu} &0.479&0.514&0.313&0.184&0.330&0.716\\
			Multi-View \cite{zhou2016multi} &0.505&0.543&0.342&0.202&0.350&0.729\\ 
			DL2R \cite{DBLP:conf/sigir/YanSW16} &0.488&0.527&0.330&0.193&0.342&0.705 \\ 
			MV-LSTM \cite{peng2016} & 0.498 & 0.538 & 0.348 &0.202&0.351&0.710 \\ 
			Match-LSTM \cite{wang2017}& 0.500& 0.537 & 0.345&0.202&0.348&0.720  \\ 
			Attentive-LSTM \cite{tan2015lstm}&0.495& 0.523 & 0.331&0.192&0.328&0.718  \\ 		
			SMN \cite{wu2016sequential}&0.529&0.569&0.397&0.233&0.396&0.724\\
			DAM \cite{DBLP:conf/acl/WuLCZDYZL18} &0.550&0.601&0.427&0.254&0.410&\textbf{0.757}\\ 
			DAM$_{single}$  &0.543&0.592&0.414&0.255&0.427&0.725\\ \midrule
			CRN + DAM$_{single}$ &0.548 &0.603 &0.428 &0.262 &0.439 &0.727 \\ 
			CRN + RL + DAM$_{single}$ &\textbf{0.552} &\textbf{0.605} &\textbf{0.431} &\textbf{0.267} &\textbf{0.445} &0.729\\
			\bottomrule
		\end{tabular}

		\caption{Evaluation results on multi-turn response selection. The numbers of baselines are copied from \cite{DBLP:conf/acl/WuLCZDYZL18} 	\label{exp:ret}
		}	 \vspace{-5mm}	
	\end{table*}
	
\begin{table}

			\centering
			\setlength{\abovecaptionskip}{0.2cm}
			\setlength{\belowcaptionskip}{-0.5cm} 
		    \small
			
			\begin{tabular}{l|c|c|c}
				\toprule
				& Win& Loss& Tie \\ \midrule
				
				CRN vs baseline &51.3\% &29.7\% &19.0\%\\
				CRN + RL vs baseline &35.7\% &21.8\% &42.5\%\\ 
				
				\bottomrule
				
			\end{tabular}	
					\caption{The result of end-to-end response selection subjective evaluation.	\label{exp:ret_human}  }	
		\end{table}
We split the last utterance from each context in the training data, and forms 0.5 million of $(q, r)$ pairs. Subsequently, we train a single-turn Deep Attention Matching Network \cite{DBLP:conf/acl/WuLCZDYZL18} consuming the pair as an input, which is denoted as DAM$_{single}$. The DAM$_{single}$ model is treated as a rank model in Section \ref{Pseudo Data Generation} and a reward function in Section \ref{selection reward}. In the testing stage, we use the CRN and the DAM$_{single}$ to assign a score for each candidate. 
Notably, the original DAM takes a context-response pair as an input, which is set as a baseline method. The parameters of the DAM is the same as its original paper. 

\subsubsection{Evaluation Results}
\label{Evaluation Results1}
Table \ref{exp:ret} shows the response selection performances of different methods. We can see that our model achieves a comparable performance with state-of-the-art DAM model, but only consuming a rewritten utterance rather than the whole context. This indicates that our model is able to recognize important content in context and generate a self-contained sentence. This argument is also verified by 1 point promotion compared with DAM$_{single}$ which only uses the last utterance as an input. Additionally, DAM$_{single}$ just underperforms DAM 1 point, meaning that the last utterance is very important for response selection. It supports our assumption that the last utterance is important which is a good prototype for context rewriting. 

\subsection{End-to-End Multi-turn Response Selection}
In practice, a retrieval-based chatbot first retrieves a number of response candidates from an index, then re-ranks the candidates with the aforementioned response selection methods. Previous works pay little attention to the retrieval stage, which just appends some keywords to the last utterance to collect candidates \cite{wu2016sequential}.

Because our model is able to rewrite context and generate a self-contained sentence. We expect it could retrieve better candidates at the first step, benefiting to the end-to-end performance. Since it is hard to evaluate the retrieval-stage, we evaluate the end-to-end response selection performance. Specifically, we first rewrite the contexts in the test set with CRN, and then retrieve 10 candidates with the rewritten context from the index\footnote{The authors \cite{wu2016sequential} share their index data with us}. DAM$_{single}$ is employed to compute relevance scores with the rewritten utterance and the candidates. The candidate with the top score is selected as the final response. The baseline model appends keywords from context to the last utterance for retrieval and use the original DAM with all context as the input to select final response.

We recruit three annotators to do a side-by-side evaluation, and the model outputs are shuffled before human evaluation. The majority of the three judgments are selected as a result. If both outputs are hard to distinguish, we choose Tie as the result.

\subsubsection{Evaluation Results}
We list the side-by-side evaluation results in Table \ref{exp:ret_human}. Human annotators prefer the outputs of our model. On account of the reranking modules are comparable, we can infer that the gain comes from the better retrieval candidates. However, reinforcement learning does not have a positive effect on this task. We find our reinforced model becomes more conservative, it tends to generate shorter rewritten utterance than our pre-training model. That may be beneficial for response re-rank, but if wrong keywords or noise words are extracted from context. It will reduce the quality of retrieved candidates, leading to an undesired end-to-end result.

\subsection{Case Study}
We list the generated examples of our models and baseline models for End-to-end Generation Chatbot and Retrieval Chatbot. Because the submission space is quite limited, we put the case study of Retrieval Chatbot in the Supplementary Material.
\begin{CJK*}{UTF8}{gbsn}
\begin{table*}[t]
			\centering
		    	\small
			\begin{tabular}{l|p{3.5cm}|p{3.5cm}|p{3.5cm}}
				\toprule
				\multirow{2}{*}{Context} &我十九号昆明飞厦门 &上了豆瓣更无聊&你5点半下班? \\
				 &I will fly from Kunming to Xiamen in 19th &Douban is so boring &Do you go off duty in 5:30?\\
				\multirow{2}{*}{Last Utterance} &我也想去&好像是这样的&是的呀，坐上班车了都 \\ 
				 &I want to go, too &Yes &Yes, I am in bus now\\ 
				 \midrule
				\multirow{2}{*}{Rewritten-Utterance} &昆明飞厦门我也想去 &豆瓣好像是这样的&是的呀，坐上班车了都下班？ \\ 
				 &I want to fly from Kunming to Xiamen, too &Yes, Douban is.&Yes, I am now in bus off duty?\\ 
				 \midrule
				\multirow{2}{*}{Rewritten-Utterance+RL} &我也想去厦门 &好像是这样的豆瓣更无聊 &是的呀5点半下班，坐上班车了都\\ 
				 &I want to go to Xiamen, too &Yes, Douban is so boring&Yes, go off duty in 5:30, I am now in bus\\ 
				 \midrule
				\multirow{2}{*}{S2SA} &那就出发吧 &你是双子&5点半 \\
				 &Let`s go &You are Gemini& 5:30\\ 
				 \midrule
				\multirow{2}{*}{HRED} &我也是  &是啊 &好吧\\
				 &Me too &Yes &Okay\\ 
				 \midrule
				\multirow{2}{*}{Our Model} &昆明大理丽江  &豆瓣毁一生 &下班了吗\\
				 &Kunming, Dali, Lijiang &Douban can ruin whole life. &Do you go off? \\ 
				 \midrule
    			\multirow{2}{*}{Our Model+RL} &厦门欢迎你  &无聊到爆&我也是坐班车 \\ 
				 &Welcome to Xiamen &I`m bored to death &I am taking the bus, too\\
				
				\bottomrule
				
			\end{tabular}	
					\caption{The examples of end-to-end response generation.	\label{exp:example_gen}  }	
		\end{table*}
\end{CJK*}
\subsubsection{End-to-end Generation Chatbots}
Table \ref{exp:example_gen} presents the generated examples of our models and baselines, our model can extract the keywords from the context which is helpful to generate an informative response, but the HRED model often generates safe responses like ``$Me too$'' or ``$Yes$''. It is because the input information from context and last utterance contain so much noise, some of the context words are useless for the last utterance to generate responses. Our model can extract important keywords from noisy context and insert them into the last utterance, it is not only easy to control and explain in a chat-bot system, but also transmit useful information directly to last utterance. The input of S2SA model is the last utterance, so it can generate diverse response due to less noise, but its relevancy with context is low. Our model succeeds fusing advantage from both models and get a significant promotion. Comparing the generated responses by our pre-training model and reinforced model, the rewritten-utterance inferred by our pre-training model may be more informative, but the final generated response may be unrelated to context and last utterance. It is because reinforcement learning can build the connection between the utterance-rewrite model with response generation model for exploring better rewritten-utterance. A better rewritten-utterance should be helpful to generate a context-related response, Too much information inserted will add noise and too little will be useless.

\section{Conclusion}
This paper investigates context modeling in open domain conversation. It proposes an unsupervised context rewriting model, which benefits to candidate retrieval and controllable conversation. Empirical results show that the rewriting contexts are similar to human references, and the rewriting process is able to improve the performance of multi-turn response selection, multi-turn response generation, and end-to-end retrieval chatbots. 

\section*{Acknowledgement}
We are thankful to Yue Liu, Sawyer Zeng and Libin Shi for their supportive work. We also gratefully thank the anonymous reviewers for their insightful comments. 

\bibliography{emnlp-ijcnlp-2019}
\bibliographystyle{acl_natbib}

\newpage
\appendix
\section{Model Details}
\subsection{Context Rewriting Model}
We employ the Adam algorithm to optimize the objective function with a batch size of 200. We set the initial learning rate as 0.0004 and reduce it by half if perplexity on validation begins to increase. Both of the encoder and decoder are 1-layer GRU, the word embedding size and hidden vector of encoder GRU are 512, the hidden size of decoder GRU is 1024. We use BPE to do Chinese word segmentation because it can solve the out-of-vocabulary word problem and reduce the size of the vocabulary. The vocabulary size of input and output are 34687. Drop out mechanism is used and equals to 0.3. We use beam search in generating rewritten last utterance and response, the beam size is always 5. For reinforcement learning, $\lambda$ is equals to 0.1.
\subsection{Single-turn Response Generation Model}
The model is pre-trained by an encoder-decoder framework with an attention mechanism, where the word embedding size is 512, the encoder is 1-layer bidirectional GRU with 1024 units in both directions, and the decoder is 1-layer unidirectional GRU with 2048 units.

\section{Case Study}
\begin{CJK*}{UTF8}{gbsn}
\begin{table*}[t]
			\centering
		    	\small
			\begin{tabular}{l|p{5.5cm}|p{5.5cm}}
				\hline
				\multirow{2}{*}{Context} &我7月20日才到奥克兰呢,你机票订了吗&那说明你很熟悉俄罗斯这个国家了啊还开贴干嘛 \\
				 &I will arrive at Auckland in July 20th, have you book the air ticket? & That means you are familiar with the country Russia, why you open new post\\
				\multirow{2}{*}{Last Utterance} &那太遗憾了我七月二十号就结束旅途了&对啊你都知道\\ 
				 &What a shame, I will finish my trip in July 20th & Yes, you have known that\\ \midrule
				\multirow{2}{*}{Rewritten-Utterance} &7月20日才到奥克兰那太遗憾了我七月二十号就结束旅途了&对啊你都知道俄罗斯这个国家了\\ 
				 &Arrive at Auckland in July 20th? What a shame, I will finish my trip in July 20th & Yes, you have known the country Russia\\ \midrule
				\multirow{2}{*}{Rewritten-Utterance-RL} &机票订了那太遗憾了,我七月二十号就结束旅途了&对啊你都知道国家了\\ 
				 &Air ticket is booked? What a shame, I will finish my trip in July 20th & Yes, you have known the country\\ \midrule
				\multirow{2}{*}{baseline} &这样啊去上学&你东北哪噶的啊\\
				 &Go to school like this? & Where is your home in northeast of China\\ \midrule
				\multirow{2}{*}{Our Model} &请问你行程规划了几天&俄罗斯不了解\\
				 &How many days have you planned your trip &I don`t know about Russia\\ \midrule
				\multirow{2}{*}{Our Model+RL} &可惜了那你还要去吗&说得容易啊\\ 
				 &What a shame, then are you going there? & Easier said\\ 
				\bottomrule
				
			\end{tabular}	
					\caption{The examples of end-to-end response selection.	\label{exp:example_ret}  }	
		\end{table*}
\end{CJK*}

\subsection{End-to-end Retrieval Chatbots}
Table \ref{exp:example_ret} presents the selected examples of our models and baselines, our model can extract the keywords from context to construct an informative utterance, which is helpful to retrieve response relative candidates from the index and select the correct response. But the baseline model retrieves candidates by concatenating extracted keywords from the context with the last utterance, the extracted keywords may be useless and even noisy, so some candidates with low correlation to utterances will be retrieved, such error will propagate to next step.

As shown in Table \ref{exp:example_ret}, after reinforcement learning, our model tends to generate shorter rewritten utterance than our pre-trained model. And the extracted keywords may be useless or lose important information. It will reduce the quality of retrieved candidates and the quality of the final selected response.

\end{document}